\documentclass[letterpaper, 10 pt, conference]{ieeeconf}  % Comment this line out if you need a4paper

\IEEEoverridecommandlockouts                              % This command is only needed if 
\usepackage{graphicx} % for pdf, bitmapped graphics files
\usepackage{amsfonts}
\usepackage{booktabs}
\usepackage{siunitx}
\usepackage{xcolor}
\usepackage{hyperref}
\title{\LARGE \bf What's in My LiDAR Odometry Toolbox? }

\author{Pierre Dellenbach$^{1,2}$, Jean-Emmanuel Deschaud$^{1}$, Bastien Jacquet$^{2}$, François Goulette$^{1}$

\thanks{$^{1}$ MINES ParisTech, PSL University, Centre for robotics, 75006 Paris, France
\{firstname.surname@mines-paristech.fr\}}

\thanks{$^{2}$ Kitware, Computer Vision Team, 69100 Villeurbanne, France, \{firstname.surname@kitware.com\}}
}

\newcommand{\tr}{ATE}

\newcommand{\tablefont}{\fontsize{7}{7}\selectfont}
\newcommand{\figvspace}{\vspace{10pt}}

\begin{document}

\maketitle
\thispagestyle{empty}
\pagestyle{empty}

%%%%%%%%%%%%%%%%%%%%%%%%%%%%%%%%%%%%%%%%%%%%%%%%%%%%%%%%%%%%%%%%%%%%%%%%%%%%%%%%
\begin{abstract}
 
 With the democratization of 3D LiDAR sensors, precise LiDAR odometries and SLAM are in high demand.
 New methods regularly appear, proposing solutions ranging from small variations in classical algorithms to radically new paradigms based on deep learning. 
 Yet it is often difficult to compare these methods, notably due to the few datasets on which the methods can be evaluated and compared. 
 Furthermore, their weaknesses are rarely examined, often letting the user discover the hard way whether a method would be appropriate for a use case.
 
 In this paper, we review and organize the main 3D LiDAR odometries into distinct categories. 
 We implemented several approaches (geometric based, deep learning based, and hybrid methods) to conduct an in-depth analysis of their strengths and weaknesses on multiple datasets, guiding the reader through the different LiDAR odometries available. 
 Implementation of the methods has been made publicly available at:
 
 \url{https://github.com/Kitware/pyLiDAR-SLAM}.

\end{abstract}

\maketitle

\section{Introduction}

{

Simultaneous Localization And Mapping (SLAM) is a fundamental building block for many robotic and vision applications.
Three-dimensional light detection and ranging (3D LiDAR) sensors that provide a highly detailed and precise geometry of the scene at high frequency, are ideally suited for this task, and the most precise SLAM systems are built on such sensors. 
The core of these methods is the LiDAR odometry which consists of predicting the relative pose between the point cloud of the new scan acquired by the sensor and the previous one, using only the data. 
Classical methods typically rely on the registration of point clouds against a map constructed from the aggregations of previous scans, and many algorithms based on this principle have been proposed.
Recently, convolutional networks were also introduced to solve this task and showed promising results. 
These methods provide different options for solving the same tasks, and we found it useful to compare them with classical approaches, aiming to help the reader answer the question : "which LiDAR odometry should I use in practice ?". 

The main contribution of this paper is the analysis of classical and deep learning-based LiDAR odometries and how they relate together, and show their respective strengths and weaknesses in a detailed comparative study realized on multiple LiDAR datasets. As secondary contributions, we offer some novel and simple design choices for implementations which are publicly available.}
 
\begin{figure}
\includegraphics[width=\linewidth]{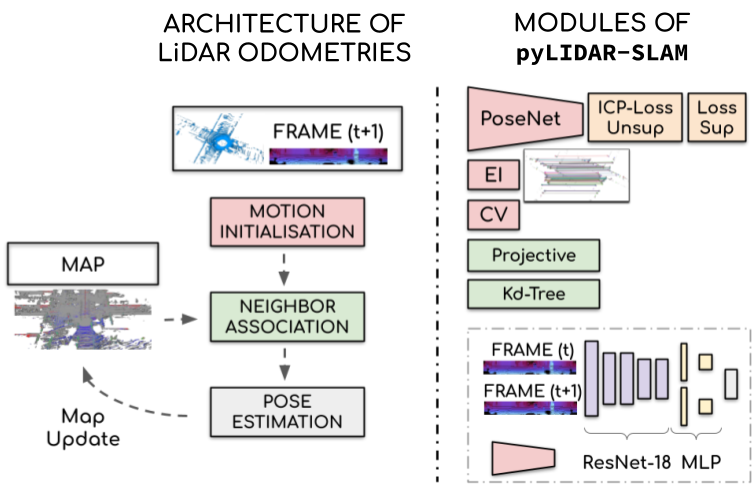}
\caption{Overview of the LiDAR odometries implemented and compared in this article. We propose a modular implementation for initialization (red) and neighbor association (green) allowing us to compare multiple approaches: deep learning, classical and hybrid methods. (EI="Elevation Image", CV = "Constant Velocity") }
\label{overview}
%\vspace{\dvs}
\end{figure}

\section{Review of State-of-the-Art LiDAR Odometries}\label{related}
In this section, we present and categorize state-of-the-art 3D LiDAR odometry methods. 
Our aim is not to be exhaustive. Instead we present and analyze the most important LiDAR odometry methods. 
We use this study to build a modular LiDAR odometry (see Figure \ref{overview} and Section \ref{methods}), allowing us to analyze the main categories of pure LiDAR odometries. 
\looseness=-1

\subsection{Classical Geometric SLAM} \label{related_classical}

Classical geometric LiDAR odometries share a common algorithmic backbone.
Point clouds are initially preprocessed, typically to select a subset of points from the scan.
The selected points are then registered into the local map, using a variant of the iterative closest point (ICP) \cite{icp} registration algorithm.
ICP requires a closest neighbor search algorithm, which must be constructed simultaneously with the map. 
It allows to build at each iteration a sum of residuals from pairs of map and scan points, which is minimized with a second-order method (Levenberg Marquardt or Gauss-Newton), updating the initial motion estimate toward the final prediction.

%TODO Mention other alignment strategies eg. NDT

\textbf{Motion initialization and preprocessing:} 
Many different methods exist to provide an initial motion estimate, from no initialization to using a simple constant velocity model (e.g. \cite{imls_slam}) or using more complex motion models (taking into account knowledge of the robot and environment). 
Other strategies take into account data from other sensors (camera \cite{vloam}, GPS, IMU \cite{locus, lio_sam}) or propose an initial frame-to-frame (\textbf{F2F}) alignment between consecutive scans \cite{loam}.
We believe that the importance of the initial motion estimate for the quality of the system is often overlooked, and we examine it in detail in Section \ref{analysis_initialization}.

The goal of preprocessing is to select good candidate points for nearest neighbor pairing. As an example, IMLS-SLAM \cite{imls_slam} removes small objects in the scan to suppress potentially mobile objects. 
Semantic segmentation can also be used to eliminate arbitrary classes (persons, bicycles, cars); however the results are not yet precise enough to improve LiDAR SLAM algorithms. Another important preprocessing step is the distortion of the frame, to compensate for the motion of the sensor during acquisition and the rolling shutter effect this creates.
Most methods use the initial motion estimate \cite{imls_slam,loam} for this first step.

\textbf{Map construction:} LOAM \cite{loam} maintains a map of features extracted from each scan and saved as a voxel grid. 
IMLS-SLAM generates a local map at each time step by aggregating the last k point clouds (k=5, 10, 20) and computing the normals of this large point cloud. 
SuMA maintains a surfel map on the GPU which is projected at each time step using OpenGL pipelines, and achieves real time. Recently, \cite{large_scale} proposed a model maintaining a truncated signed distance function (TSDF) on a sparse voxel grid using voxel hashing \cite{voxel_hashing} for a scalable and GPU-friendly model. 

%\begin{figure}
%\includegraphics[width=\linewidth]{images/map_aggregated}
%\caption{The local map (gray) as an aggregation of k=20 consecutive point clouds. Points in the map for which the normals were computed are colored according their normals.}
%\label{aggregated_map}
%\vspace{\dvs}
%\end{figure}

\textbf{Neighbor association:} The two dominant approaches are projective neighborhood association \cite{large_scale, suma}, or tree-based neighbor search \cite{imls_slam,locus,loam}. 
The former is typically addressed by projecting the new target scan and the map into the image plane using a spherical projection (in the coordinate system of the map or target) which is typically parallelizable on the GPU and allowing real-time odometry. 
Tree-based search, allows more precise neighborhood computation (for neighbor search and neighborhood feature estimation, like normals), but the search and construction are often more costly in time. 
Real time is achievable, with optimized implementations, and/or by reducing the number of points in the tree and the number of queries. But using all the information in the frame is a challenge for these methods, notably when considering recent LiDAR with 64 to 256 laser channels, which typically produce more than a million points per seconds. 
%Voxel hashing \cite{voxel_hashing} proposes a new option for approximate neighbor association, but has not been explored yet in traditional LiDAR odometry methods.

\textbf{Energy minimization:} The standard point-to-point distance minimized by ICP has a very small convergence domain, outside of which it tends to fall into a bad local minimum. 
The most popular metric is the point-to-plane distance, which takes into account the neighborhood structure and allow points to slide on the local plane tangent to the neighbor.
Increasingly popular is generalized ICP \cite{generalized_icp}  which uses the covariance of the reference points to construct the residuals, which was chosen by \cite{locus} for their LiDAR odometry.
Note that both methods require the computation/estimation of a covariance on the map, and thus, require a neighborhood structure.
Despite these better objectives, the convergence domain of ICP remains small, and successful registration highly depends on a good initial estimate pushing the target scan into it.

\subsection{Deep LiDAR Odometries}

Recently, several methods were proposed to regress the parameters of the relative pose between two consecutive frames. 
Many convolutional network architectures were proposed for this task (for simplicity we name all such networks PoseNet). 
DeepLO \cite{deeplo} extracts features from a spherical projection of point clouds (named Vertex Map, as they preserve the x,y,z coordinates by contrast with a range image) and their associated normal maps. 
Others combine features from different sources or sensors. \cite{visual_deeplo} uses a Siamese pyramid network which combines features from two pairs of sparse depth and RGB images and regresses the relative motion, and depth. 

%Or intermediary predictions as input :
%\cite{two_stream} combines two streams of features, one from a pair of depth predictions and the other from a pair of images, to produce a single pose estimation between the two frames. 

\textbf{Supervised:} LO-Net \cite{lo_net} is the first method to use consecutive LiDAR frames as input for PoseNet, to regress a relative transform. 
They train PoseNet on KITTI \cite{kitti} dataset using the ground truth for supervision. 
Typical of supervised methods on such datasets, their results show signs of overfitting with a large discrepancy on results from the training and test sets.

\textbf{Unsupervised:}
DeepLO \cite{deeplo} introduces a self-supervised training strategy using point-to-plane error coupled with projective  neighbor  association.  
This  approach  allows  PoseNet to  learn LiDAR  odometry  without  the  expensive  cost  of  generating accurate ground  truth  labels  and  was  refined in \cite{eth_deeplo}.  
We  present and implement this approach in depth in Section \ref{methods},  proposing  an improved  training strategy. 
Similarly to supervised methods, these approaches typically show overfitting, and perform much worse on test data than on training data. 
However, and in contrast to supervised approaches, unsupervised methods allow to construct an estimate of the trajectory only from the data.

\subsection{Hybrid LiDAR Odometries}

Despite promising results for these new deep learning methods, there remains a large performance gap between classical and deep LiDAR odometries. 
However, both approaches use radically different and potentially complementary mechanisms to predict the relative pose, motivating different kinds of hybrid pipelines, which we present below. 

\textbf{PoseNet as initialization before mapping:} As mentioned in Section \ref{related_classical} the performance of classical methods strongly depends on good motion initialization before registration, and PoseNet is proposed as a tool to that end. LO-Net \cite{lo_net} uses PoseNet as the initial estimate before the mapping stage, trained using supervision from the ground truth. 
Similarly, \cite{eth_deeplo} uses LOAM's mapping to refine their PoseNet's prediction, showing that it outperforms the default LOAM on multiple datasets. 
However, work remains to be done to prove PoseNet's practical interest over standard initialization strategies, and we investigate this in more detail in Section \ref{analysis_initialization}.  

\textbf{Learning methods to improve neighbor association:} DMLO \cite{dmlo} proposes a novel hybrid strategy which learns to construct direct and non-iterative registration between consecutive scans. 
Their approach consists of learning to predict a neighbor association with a confidence score which allows them to robustly select the best matching pairs of points, before directly minimizing their point-to-point distance with singular value decomposition, thus avoiding the iterative refinement of the ICP. 
However, this method requires ground truth labels for a supervised training scheme.

\section{Details of the Methods Implemented and Datasets Considered}
\label{methods}

Following the previous taxonomy, we present the algorithms implemented for our analysis in Section \ref{analysis}, as well as the datasets considered. 

\subsection{Classical registration-based SLAM}

\begin{figure}
\figvspace
\centering
\includegraphics[width=0.8\linewidth]{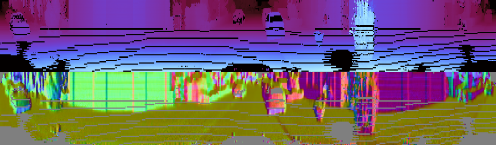}
\caption{Example of a new frame projected into a range image (top) and normal map (bottom) from KITTI dataset \cite{kitti}.}
\label{vertex_map}
%\vspace{-10}
\end{figure}

We designed our classical odometries to be simple and modular, allowing us to test and compare different strategies for each module. 
Our system uses a standard frame-to-model (\textbf{F2M}) registration method based on the ICP \cite{icp} algorithm. 
At each iteration of the registration algorithm, a set of point correspondences between a new frame (see figure \ref{vertex_map}) and the model is selected, which is used to update the relative motion estimate and move the new frame before the next iteration. 
The details are presented below.

\textbf{Energy minimized:}
Following \cite{imls_slam} and \cite{suma}, the relative pose between two time steps $\mathbf{\delta T} \in SE(3)$ is estimated by minimizing the point-to-plane distance between neighbor points.  
Given a set of pair of points and normals $(\mathbf{p}_i, \mathbf{q}_i, \mathbf{n}_i)_{i\in[1,n]}$, where $\mathbf{p}_i$ are the points of the new scan, and $\mathbf{q}_i, \mathbf{n}_i$ the points and normals of the reference map, the point-to-plane objective is: 
%\vspace{-2}

$$E(\mathbf{\delta T}) = \sum_{i=1}^n{ w_i \cdot ((\mathbf{\delta T} * \mathbf{p}_i - \mathbf{q}_i) \cdot{} \mathbf{n}_i})^2 $$
%\vspace{-2}

Minimizing the objective using second-order methods (Gauss-Newton), leads to fast convergence, but is highly sensitive to outliers or imprecise neighbor pairing. IMLS-SLAM \cite{imls_slam} eliminates residuals greater than 0.5m, and \cite{suma} minimizes a robust Huber loss.
We found that using the weighting scheme $w_i = \text{exp}(- (\mathbf{p}_i - \mathbf{q}_i)^2/\sigma^2)$, which gives more importance to pairs the closer they are, yields (marginally) the best results. 
We set $\sigma$ to 0.5m in all the experiments.

\textbf{Local map and neighborhood:}
For the classical frame-to-model LiDAR odometry, and to investigate both projective (\textbf{P-F2M}) and KdTree-based (\textbf{Kd-F2M}) neighbor association, we propose two types of models.
The first model consists of aggregated point clouds similarly to \cite{imls_slam}. 
The previous $k=30$ registered point clouds are kept in memory, and a KdTree is constructed at each time step, allowing pairing points from the new scan and the map, and building the corresponding normals for the local map. 
Computing a KdTree at each iteration is slow due to the large number of points returned by the LiDAR sensor. 
This can be addressed by reducing the number of queries constructed \cite{imls_slam} or using faster approximate data structures (e.g. voxel hashing \cite{voxel_hashing}). 

A different nature of neighbor association is obtained by projection in the image plane much like SuMA \cite{suma}, and its main interest lies in its capacity to be accelerated on the GPU. 
Similarly to \cite{suma}, we use a projection of the point cloud into a spherical image (named vertex map) which stores in each activated pixel (for which there exists a projected point) the three coordinates $x,y,z$ (see Figure \ref{vertex_map}). 
Each new scan is projected into a vertex map and the normals are then computed using activated neighbor pixels \cite{fast_normals}.

Similarly to our KdTree-based implementation, the local map consists of the $k=30$ previous frames. 
At each time step, each point cloud of the map is transformed in the reference frame of the last registered scan and then projected into a vertex map. 
Each point of the new scan is associated with its closest neighbor among points of the projected map sharing the same pixel values. 
We chose this scheme as it mitigates the problem of 2D projective data association, such as occlusions and mobile objects polluting the map.

\begin{figure}
\figvspace
\centering
\includegraphics[width=0.85\linewidth]{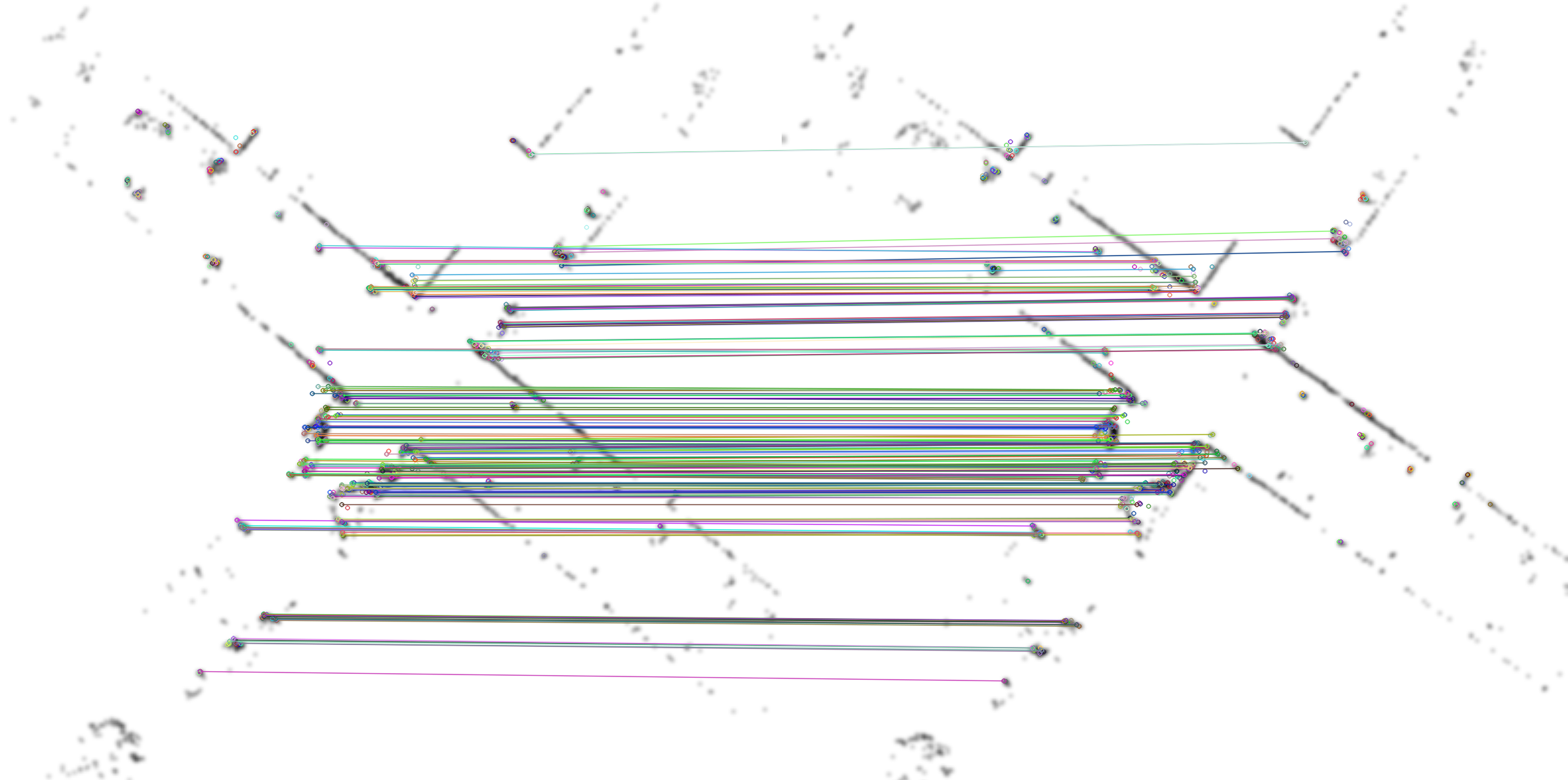}
\caption{Initialization strategy of two consecutive frames using elevation images (EI): Feature points are detected and robustly matched between the two images by fitting an  homography between consecutive elevation images.}
\label{orb_matching}
%\vspace{-5}

\end{figure}

\textbf{Initialization:}
As mentioned in Section \ref{related_classical} ICP has a small convergence domain, and optimal registration requires a good initial prediction.
When the moving object has strong inertia (relative to the scan acquisition frequency), for example, in driving scenarios, the simple constant velocity (\textbf{CV}) approximation consisting of using the previous relative motion estimated is often a correct initial estimate (cf \cite{imls_slam,suma}). 
Additionaly, and to examine the impact of initialization on the performance of LiDAR odometries, we experiment with PoseNet as motion initialization before \textbf{F2M} alignment.
For many robotic applications, the motion of the sensor is essentially 2D, and estimating this 2D motion often provides a good initial estimate. 
Thus, we propose a simple and fast 2D feature-based registration mechanism to provide this estimate.
For each new point cloud, an elevation image (\textbf{EI}) is built from points above the sensor (with a pixel size of 30 cm, and a resolution of $800\times 800$), and ORB feature points \cite{orb} are extracted. 
The motion is estimated by robustly fitting an image homography between two consecutive sets of feature points (see Figure \ref{orb_matching}), setting $N_I=100$ inliers as a threshold for a correct alignment.
These three initialization strategies (PoseNet, EI, CV) are studied in detail in Section \ref{analysis_initialization}.

\subsection{Deep LiDAR Odometries}

We also propose a deep LiDAR odometry using ResNet-18 based PoseNet similar to \cite{deeplo}. 
Two consecutive vertex maps (VMaps) are concatenated in the image channel dimension, and PoseNet outputs the six parameters of the relative transform between the two scans (using Euler angles for the rotation parameters). 

\textbf{Unsupervised training:}
Similarly to \cite{deeplo, eth_deeplo}, we use a self-supervised objective to train PoseNet using a projective point-to-plane geometric alignment loss. 
Given a target and a reference VMap, a reference normal map (NMap) and the estimated relative pose $\delta \mathbf{T}$ from PoseNet, the target point cloud is transformed using $\delta \mathbf{T}$ and reprojected in the reference frame. 
For each of the $n_k$ pixels activated in the reference and the new target VMap, a point-to-plane residual is built leading to the following loss:  
$$ L_{icp}(\delta \mathbf{T}) = \frac{1}{n_k}\sum_{i=1}^{n_k} \rho ( r_i )$$
The loss proposed by \cite{deeplo} is unstable, and an out-of-field-of-view loss was proposed to make training possible. 
In contrast \cite{eth_deeplo} proposes to use a KdTree during training, for neighbor association instead, allowing them to filter outliers using a radius search. 
However, we found that projective association during training is still feasible and stable and yields results comparable to \cite{eth_deeplo}. 
Key to achieving this is selecting only activated pixels, and using a loss robust to outliers. Using for residuals $\rho(r_i)=|\mathbf{(\delta T * p_i - q_i)\cdot n_i}|$ allows stable training; however, we observe better results using a weighted $L_2$ loss: $\rho(r_i)=w_i \cdot |  (\mathbf{\delta T * p_i - q_i) \cdot n_i}|^2$ similarly to the \textbf{F2M} energy, which suppresses the effect of bad associations. 
Unlike \cite{eth_deeplo}, our method does not require precomputation of the normals for each scan before training, or the use a KdTree, allowing faster training.

\textbf{Supervised training:} Our main focus is on the use of unsupervised LiDAR odometries. However, to examine PoseNet generalization capacity on the challenging dataset NCLT (see Section \ref{datasets}) where despite our best efforts, unsupervised training of PoseNet has failed, we also propose supervised training of PoseNet similar to LO-Net \cite{lo_net}. 
%Given a ground truth $\mathbf{x}_{gt}=(\mathbf{t}_{gt},\mathbf{e}_{gt}) \in \mathbb{R}^6$, and a PoseNet prediction $\mathbf{x}_{pred}=(\mathbf{t}_{pred},\mathbf{e}_{pred}) \in \mathbb{R}^6$, the supervised loss is 

%$$L_{sup} = ||\mathbf{t}_{gt} - \mathbf{t}_{pred} ||^2_2 \cdot{} e ^{s_{1}} + ||\mathbf{e}_{gt} - \mathbf{e}_{pred} ||^2_2 \cdot{} e^{s_{2}} - (s_{1} + s_{2})$$

\subsection{Datasets and Metrics} \label{datasets}

 Analyzing LiDAR odometries is a challenging task, notably due to the few publicly available and annotated LiDAR datasets. 
 We now present the datasets and metrics we found most relevant for this study ({\bf\texttt{pyLIDAR-SLAM}} incorporates more datasets not mentioned here due to the lack of space).

\textbf{KITTI \cite{kitti}}
is the reference dataset for LiDAR odometries in driving scenarios, and the odometry split consists of 21 suburban, highway, or city driving sequences of point clouds captured with a Velodyne HDL-64 LiDAR sensor rotating at 10Hz, 
11 of which possess a ground truth. 
%However many recent LiDAR odometries perform so well on KITTI that it becomes complicated to separate them. 
%For Deep Learning odometries, the small size of KITTI's annotated data for the odometry split ($\sim$20000 training scans), limits the impact of experimental studies. 
%Essentially, PoseNet is trained on sequences \texttt{00-08} of the split, and tested on sequences \texttt{09-10} which are two small, unchallenging sequences (few rapid turns, few mobile objects, environments full of geometric features).
KITTI is the principal dataset for the present studies, however, and in contrast to existing approaches, we make the most of the dataset by investigating different training splits, to find challenging scenarios for our algorithms, allowing us to draw meaningful conclusions in Section \ref{analysis}. We use principally KITTI's metric Average Translation Error (\tr), which quantifies the average translation error over segment of different lengths from 100 m to 800 m.

\textbf{Ford Campus \cite{ford_campus}} is another driving dataset consisting of two sequences, of another HDL-64 for a LiDAR rotating at 10 Hz representing 9996 frames, on the Ford Campus of Dearborn, Michigan. 
The motion of the car in this dataset is much simpler than for KITTI (many fewer turns, long stops at red lights, long straight lines). However, it offers new environments (parking, small American town), as well as other challenging elements (e.g. sequences with many mobile objects). 
Furthermore, the same LiDAR as KITTI is used, in the same configuration, allowing meaningful complementary analysis. \looseness=-1

% TODO : MICHIGAN
\textbf{NCLT \cite{nclt}} is a large dataset designed for long-term vision and LiDAR tasks. 
A rotating Velodyne HDL-32 (10Hz) mounted on a Segway makes multiple runs through Michigan University's North Campus, over the course of a year. 
Each sequence is composed of $\sim$20000 motion-compensated LiDAR frames and proposes a new trajectory around the campus. 
Despite the slow speed of the robot, the dataset is challenging for pure LiDAR odometry due to the fast rotations induced by the motion of the Segway. 
Thus, despite a motion which is principally planar, each frame is hard to register on the previous one, which is ideal for testing odometries. 
Furthermore, the large quantity of available data in a shared environment makes it an ideal candidate for evaluating the generalization of training tasks. 
We used sequence \texttt{01/08} for testing and 10 other sequences for training (captured  between \texttt{01/22} and \texttt{03/31} in 2012).

%\begin{figure}
%\includegraphics[width=\linewidth]{images/michigan_overlaid.png}
%\caption{Trajectories of the sequences of the NCLT dataset considered in this article, overlaid on top of each other. The average trajectory length is 6.1km for an average of 26000 LiDAR frames.}
%\label{michigan_overlaid}
%\end{figure}

\subsection{Results and Comparison with State-of-the-art LiDAR Odometries} \label{methods_results}

% TODO : INTERPRET RESULTS MORE IN DEPTH

\begin{table*}
\figvspace
\centering

\begin{tabular}{l|*{10}{c}c} 

 \multicolumn{1}{l|}{} & \multicolumn{1}{c|}{\texttt{00}*} & \multicolumn{1}{c|}{\texttt{01}*} & \multicolumn{1}{c|}{\texttt{02}*} &  \multicolumn{1}{c|}{\texttt{03}*} & \multicolumn{1}{c|}{\texttt{04}*} & \multicolumn{1}{c|}{\texttt{05}*} & \multicolumn{1}{c|}{\texttt{06}*} & \multicolumn{1}{c|}{\texttt{07}*} & \multicolumn{1}{c|}{\texttt{08}*} & \multicolumn{1}{c|}{\texttt{09}} & \multicolumn{1}{c}{\texttt{10}}  \\
\toprule
\toprule
\textbf{Classical Tree-Based LiDAR odometries} \\
\midrule

\multicolumn{1}{l|}{IMLS-SLAM \cite{imls_slam}} & \textbf{0.5} & 0.82 & 0.53 & 0.68 & \textbf{0.33} & 0.32 & \textbf{0.22} & 0.33 & 0.8 & 0.55 & 0.53 \\
LOAM \cite{loam} & 0.78 & 1.43 & 0.92 & 0.86 & 0.71 & 0.57 & 0.65 & 0.63 & 1.12 & 0.77 & 0.79\\
\midrule 
\textit{EI + Kd-F2M} & 0.53 &	0.79 &	0.52 & 0.69 & 0.45 & 0.34	& 0.31 & 0.36 & 0.79 & 0.54 & \textbf{0.51}  \\
\textit{CV + Kd-F2M} & 0.51 &	\textbf{0.79} & \textbf{0.51} & \textbf{0.64} &0.36 &	\textbf{0.29} & 0.29 & \textbf{0.32} & \textbf{0.78} & \textbf{0.46} & 0.57 \\
\midrule
\midrule
\textbf{Projective LiDAR Odometries} \\
\midrule

\multicolumn{1}{l|}{SuMA \cite{suma}} & 0.68 & 1.70 & 1.20 & \textbf{0.74} & \textbf{0.44} & 0. 43& 0.54 & 0.74 & 1.20 & 0.62 & \textbf{0.72} \\ 
\midrule 
\textit{EI + P-F2M}  &\textbf{0,57}& \textbf{0.67}&	\textbf{0.62}&	0.83&0.51 &	\textbf{0.37}&	\textbf{0.36}&	\textbf{0.32}&	\textbf{0.93}& \textbf{0.60} &	1.01 \\
\textit{CV + P-F2M}  & 0.57 & 0.71 & 0.62 &	0.82& 1.12&	0.37&	0.37&	0.33&	0.93&	0.61&	1,01 \\

\midrule
\midrule
\textbf{Unsupervised Deep LiDAR Odometries} \\
\midrule
DeepLO \cite{deeplo} & 1.90 & 37.83 & 2.05 & 2.85 & 1.54 & 1.72 &  \textbf{0.84} & \textbf{0.70} & \textbf{1.81} & 6.55 & 7.74 \\
Self-supervised LO \cite{eth_deeplo} & - & - & - & - & - & - & - & - & - & \textbf{6.05} & \textbf{6.44} \\
\midrule  

\textit{PoseNet + ICP loss} & \textbf{1.33}&	7.11&	1.81&	3.09&	0.93&	1.38&	1.43&	0.84&	2.27&	6.31&	8.99\\
\textit{PoseNet + Weighted ICP loss} &1.36&	\textbf{1.88}&	\textbf{1.46}&	\textbf{2.33}&	\textbf{0,969}&	\textbf{1.26}&	1.05 &	1.07&	2.05&	6.79&	7.6 \\

\midrule 
\midrule
\textbf{Hybrid LiDAR Odometries} \\
\midrule
LO-Net \cite{lo_net} & 0.78 & 1.42 & 1.01 & \textbf{0.73} & 0.56 & 0.62 & 0.55 & 0.56 & 1.08 & 0.77 & 0.92 \\
\midrule
\textit{PoseNet + P-F2M} &0.601 & \textbf{0.67} &	0.68&	0.85&	0.56&	0.41&	0.39&	0.34&	0.96&	0.69&	1.07 \\
\textit{PoseNet + Kd-F2M} & \textbf{0.55} & 0.85 & \textbf{0.58} & 0.74 & \textbf{0.44} & \textbf{0.34} & \textbf{0.36} & \textbf{0.33} & \textbf{0.88} & \textbf{0.61} & \textbf{0.83} \\

\bottomrule
\end{tabular}
\caption{\tablefont Average Translation Error (ATE) as a percentage of distance traveled for KITTI's Dataset, comparing Published LiDAR Odometries and the Proposed Methods (italic)
 / EI="Elevation Image", CV="Constant Velocity", F2M="Frame-to-Model", * are training sequences for PoseNet}
\label{main_table_kitti}
\end{table*}

Algorithms presented in Section \ref{methods} were implemented in Python and PyTorch. All networks were trained using Adam with default parameters, for 100 epochs, with an initial learning rate of $10^{-4}$ divided by two every 20 epochs. 
For the classical odometry, the local map is built from the last 30 pointclouds registered. To limit the number of points in the \textbf{Kd-F2M} a grid sampling is performed with a voxel size of 0.4m, and the dimension of the vertex map for \textbf{P-F2M} (and during the training of PoseNet) is 64$\times$720 (this resolution acts as a spherical sampler and reduces the inbalance of point densities).

To demonstrate that the implementations are comparable to state-of-the-art methods, we ran experiments on KITTI \cite{kitti} and present the results in Table \ref{main_table_kitti} ($*$ denotes sequences used to train PoseNet).
Our \textbf{Kd-F2M} runs at 2 Hz on KITTI and our \textbf{P-F2M} at 10 Hz, on a i7-9750H CPU @ 2.60GHz with a GeForce RTX 2080 GPU. 
Table \ref{main_table_kitti} shows that we obtain near state-of-the-art results for our tree-based LiDAR odometries (Kd-F2M) and even slightly better results than the published methods in our projective method (P-F2M), in our PoseNet version with weighted L2 Loss and in the hybrid LiDAR odometry. These results legitimize the different choices we made before we launch a more constructive analysis in section \ref{analysis} of the different approaches in LiDAR SLAM based on our implementations.

More interestingly, we can already see from table \ref{main_table_kitti} that tree-based methods generally obtain better results than projective methods.
This is natural, as tree-based method offer more precise neighbor associations, at the cost of slower execution. 
However, the gap is not very important (the \textbf{P-F2M} implementation obtains better results than LOAM).
In addition, we can see that PoseNet as an initialization strategy is (marginally) better than other strategies; we examine this observation in more detail in Section \ref{analysis_initialization}.
Finally, and perhaps most important, there is still a large gap between the classical \textbf{F2M} LiDAR odometries and unsupervised deep learning based odometries.
One goal of this article is to answer the following question: Despite this gap, can deep LiDAR odometries become practically useful and complement classical LiDAR odometries?
We address this point in the next section.\looseness=-1

\section{Comparative Study}
\label{analysis}

In this section, we evaluate the algorithms proposed in Section \ref{methods} in a more challenging setting and present the strengths and weaknesses of the proposed LiDAR odometry methods.
This in turn allows us to understand how and when these odometries complement each other.
More precisely, we inspect the claim that PoseNet outperforms frame-to-frame (\textbf{F2F}) ICP-based odometries in Section \ref{posenet_vs_frame_to_frame}, and show that it takes little effort to build a classic \textbf{F2F} odometry contradicting this claim.

In Section \ref{understanding}, we take a closer look at PoseNet and show the limits of its generalization capacity, showing that further research is needed before practical use.
In Section \ref{analysis_initialization}, we examine different initialization to classical \textbf{F2M} odometries and show that PoseNet provides marginal to no improvements over simpler strategies, depending on the training data considered. 
Finally, section \ref{limits_geometric}, we expose the weaknesses and failures of classical LiDAR odometries, thus revealing scenarios where PoseNet can already be useful.

\subsection{PoseNet vs. Classical Frame-to-Frame Odometry} \label{posenet_vs_frame_to_frame}

% A REECRIRE QUAND LES RESULTATS SONT LA
%
% GOAL OF THIS SECTION : DEBUNK THE MYTH THAT Posenet is better than  FRAME-TO-FRAME ICP based odometry  REGISTRATION IS BAD
% SHOW THAT WITH LITTLE EFFORT, FRAME-TO-FRAME ICP CAN BE MADE BETTER THAN POSENET
%
% EXPLAIN THAT IT IS NORMAL : FRAME-TO-FRAME ICP HAS THE SAME SUPERVISION SIGNAL THAN POSENET + ICP has iterative refinment method
% EXPLAIN THE DIFFERENT MACHINERIES

\begin{table}
\centering

\begin{tabular}{l|c|c}
& \texttt{00-08}* & \texttt{09-10} \\
\toprule
\toprule
PoseNet (ICP loss) & 2.24 & 7.65\\ \hline
PoseNet (Weighted ICP loss) & 1.49 & 7.19\\ \hline
NI + P-F2F & 40.1 & 30.4\\ \hline
CV + P-F2F & 1.46 & 1.7\\ \hline
EI + P-F2F & 1.47 & 1.9 \\\hline
NI + Kd-F2F & 24.18 & 14.04 \\ \hline
CV + Kd-F2F & 1.41 & 1.84 \\ \hline
EI + Kd-F2F & 1.41 & 1.87  \\
\bottomrule

\end{tabular}
\caption{\tablefont  Absolute Translation Error (ATE) on KITTI: compares PoseNet to different Frame-to-Frame (F2F) methods / NI ="No Initialization"}
\label{table_frame_to_frame}
%\vspace{\dvs}
\end{table}
PoseNet is often compared to the default \textbf{F2F} ICP registration (using point-to-plane or point-to-point error metric) \cite{deeplo, eth_deeplo}. 
However, we found the comparison unfair, as little effort is required to build a decent frame-to-frame LiDAR odometry which obtains comparable and even slightly better precision on the training and test sets. 
Table \ref{table_frame_to_frame} confirms that using a naïve point-to-plane registration with no initialization (\textbf{NI}) leads to worse performances than PoseNet (rows 3 and 6 of the table).  

However, using simple initialization methods (\textbf{CV, EI}), is enough to outperform PoseNet (rows 4 and 5).
This shows that frame-to-frame can be made much more robust and precise than advertised. 
This is reassuring as \textbf{P-F2F} alignment is precisely the supervision signal used to train PoseNet. 
Furthermore, PoseNet and geometric LiDAR odometries use very different mechanisms to predict the motion. 
ICP-based methods allow for iterative refinement of an estimate. 
In contrast PoseNet maps features extracted from the pair of scans to a relative pose, via spaces deformed by training data. 
Thus, the precision of the latter mechanism is limited by the representative power of the network and depends notably on the number of parameters and the volume of training data, whereas the former solely depends on noise and the ability to make correct neighbor associations. 

Note that the results in Table \ref{table_frame_to_frame} hint at the practical utility of PoseNet: When the motion cannot be properly initialized, either due to abrupt motion or when initializing the map, PoseNet can use all the data as a regularizer and obtains decent predictions, outperforming standard \textbf{F2F} methods with no initalization. 
However, PoseNet requires a supervision signal, which in self-supervised settings is based on classical methods. 
Thus we have the following paradox: when this signal is possible, classical methods typically perform well (as demonstrated on KITTI).
More insight into this subject is provided in Section \ref{understanding}, by examining the generalization capacity of PoseNet.

% TODO : If time train PoseNet on Michigan with EI loss and compare to frame to frame / frame to frame 
% ESSENTIALLY POSENET Better when it can : THIS IS NOT PROVEN IN THE CONTEXT OF DRIVING ON KITTI (when simple CV)
%

\subsection{Understanding the Limitations of PoseNet for RPR} \label{understanding}

PoseNet for Relative Pose Regression (\textbf{RPR}) makes the false promise of learning to find the optimal pose between consecutive LiDAR frames. \cite{understanding} showed the limits of convolutional networks for absolute pose regression, showing notably their behavior is close to image retrieval pipelines, without an understanding of the geometry of the scenes, which leads to poor generalization capacities. In this section, we show that this is also true for PoseNet for \textbf{RPR} and explore the limits of PoseNet generalization to unknown motion (label distribution) or unknown environments (input distribution).\looseness=-1

\begin{table}
\figvspace
\centering
\begin{tabular}{l|c|c} 
& \{\texttt{00, 02-10}\} & \texttt{01} \\
\toprule
\toprule
PoseNet \{\texttt{00, 02-10}\}* &  1.51 &  56.04 \\ \hline
PoseNet \{\texttt{00, 02-21}\}* & 1.63 & 13.46 \\ \hline
CV + P-F2F & 1.52 & 2.17 \\
\bottomrule
\end{tabular}
\caption{\tablefont Average Translation Error on sequence \texttt{01} of KITTI  shows that PoseNet fails to infer unobserved motion during training / CV="Constant Velocity", P-F2F="Projective Frame-to-frame"}
\label{generalisation_motion_table}
\end{table}

\begin{figure}
% Plot showing the distribution of motion 

\centering
\includegraphics[width=0.85\linewidth]{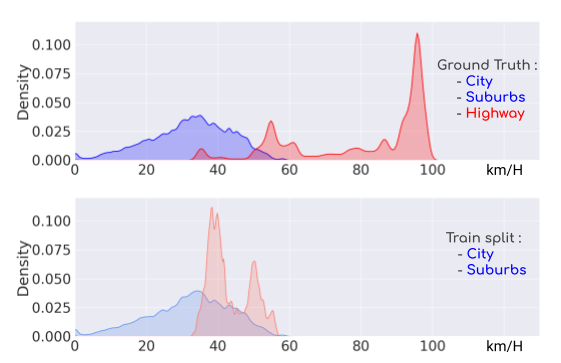}

\caption{Distribution of the speed (scaled in km/h) for KITTI's agglomeration sequences (\texttt{\{00, 02-10\}} in blue) and highway (\texttt{01} in red) computed from the ground truth and from PoseNet's prediction (light blue and red) for the training split \texttt{\{00, 02-10\}} (bottom). The PoseNet incapacity to generalize well to motion unobserved during training is demonstrated.}
\label{distribution_motion}
%\vspace{-5}
\end{figure}

\textbf{Generalization to unobserved motion:} Tables \ref{main_table_kitti} and \ref{table_frame_to_frame} show the discrepancy in performances between training and testing for PoseNet. 
To complement this observation, we examine the behavior of PoseNet on motion unobserved during training.
KITTI's sequence \texttt{01} (highway) is unique among the dataset which is mainly composed of residential and city environments in the sequences \{\texttt{00, 02-10}\}. The test split \{\texttt{11-21}\} also contains several road sequences but with a maximum speed of 90 km/h.
Figure \ref{distribution_motion} presents the distribution of the motion from the ground truth and PoseNet predictions trained on the split \{\texttt{00, 02-10}\} (city, suburbs). We also trained PoseNet on the split \{\texttt{00, 02-21}\} (city, suburbs, roads), and Table 
\ref{generalisation_motion_table} shows the trajectory errors for both splits.
Both trainings demonstrate the poor generalization capacity of PoseNet to the unobserved motion of sequence \texttt{01}.
The scale of the motion predicted by PoseNet does not leave the distribution observed during training (as seen in Figure \ref{distribution_motion}). 
Interestingly, when roads are added to the training set, PoseNet can relate the motion of the sequence \texttt{01} to the motion of the roads (with speed greater than 80 km/h), however is unable to interpolate beyond. 
This illustrates the same behavior as for absolute pose regression: at best PoseNet associates a pair of frames with the closest motion observed in the dataset. 

%We trained PoseNet on all the remaining data of the odometry benchmark dataset, and show the results of PoseNet odometry in table \ref{generalisation_motion_table}. 
%PoseNet performs much better on the training set than the test set as previously observed, more importantly its performance on sequence \texttt{01} is much worse than on the test sequences \texttt{09-10} when training on the default split \texttt{00-08}, despite using more than twice as much data.
% Figure 4 shows that PoseNet is unble to predict motion outside of the distribution of its training data, thus .
%This confirms the insight of \cite{sattler2019understanding}, that current PoseNet models typically behave poorly when required to predict new, unobserved relative motion. 

%{\color{red} \bf TODO : Redo above to take into account the two distributions + table \ref{table_michigan_initialization}}

% TODO Show the prediction of PoseNet on sequence 01 to see the distribution of the length (relevant if it is contained in the blue distribution)
% Test PoseNet generalization capacity on much more data

\textbf{Generalization between datasets} We now examine whether PoseNet can generalize between comparable datasets.
The answer is essentially not yet. We trained PoseNet on KITTI and tested on Ford Campus dataset (and reciprocally). We present the results in table \ref{generalization_ford_campus}. 
PoseNet showed a good performance on the training dataset, however obtains poor results on the other dataset.
This is underwhelming as both datasets were acquired using the same LiDAR in the same configuration, and PoseNet was trained using the same spherical projection. 
It shows the limited generalization capacities of PoseNet to different acquisitions in comparable scenarios, and that further work is required to achieve that goal which could unlock further practical interests for PoseNet. 
 
% ////////////////////////////////////////////////////////
% TODO ONLY IF INTERESTING RESULTS WITH SUPERVISED PoseNet
% ////////////////////////////////////////////////////////
%\textbf{Generalization to a large volume of Data}
%PoseNet trained with supervision on Michigan (using a large number of sequences for training) / Studying the volume of the data required to have.
%\begin{table}
%\centering
%\begin{tabular}{l|c|c} 
%& 01-08 & Train average \\
%\toprule
%\toprule
%PoseNet (sup 3 sequences) & {\color{red} \textbf{TODO}}  &  \\ \hline
%PoseNet (sup 5 sequences) &  & \\ \hline
%PoseNet (sup 10 sequences) &  &  \\ \hline
%PoseNet (unsup 10 sequences) & & \\ \hline
%EI + Kd-F2F & & - \\ \hline
%EI + Kd-F2M & & - \\ \hline
%\bottomrule
%\end{tabular}
%\caption{\tr \hspace{} for PoseNet trained on multiple data splits. {\color{red}\bf TODO : Only add if it makes sense, and show the metric's dependance to the size of the training data }}
%\label{table_michigan_sup}
%\end{table}

\begin{table}
\figvspace
\centering
\begin{tabular}{l|c|c} 
& Ford Campus & KITTI (\texttt{00-10}) \\
\toprule
\toprule
PoseNet (Ford Campus)* & 3.94  & 73.42 \\ \hline
PoseNet (KITTI \texttt{00-21})* & 71.56 & 1.54 \\ \hline
CV + P-F2F & 4.32 & 1.52\\ \hline
CV + P-F2M & 2.04 & 0.69 \\ 
\bottomrule
\end{tabular}
\caption{\tablefont Average Translation Error (\tr) shows Poor Generalization of PoseNet between KITTI and Ford Campus datasets}
\label{generalization_ford_campus}
%\vspace{-20}
\end{table}

\subsection{Initialization Strategies} \label{analysis_initialization} 
% GOAL OF THIS SECTION : SHOW THAT POSENET IS RARELY BETTER INITIALIZATION STRATEGY THAN DEFAULT ONE
%

As mentioned in Section \ref{related} PoseNet for LiDAR-based \textbf{RPR} has originally been introduced by LO-Net \cite{lo_net} as a supervised network predicting a motion estimate before mapping.
This point of view was extended in \cite{eth_deeplo} which provides unsupervised training for PoseNet. 
This is a theoretically sound approach, as initialization is critical for successful \textbf{F2M} odometry, due to the small convergence domain of ICP-based registration. 
However, for a more complete analysis, we need to compare it to simpler approaches, and in different and more challenging settings which we do in this section, using the algorithms presented in Section \ref{methods}.

\textbf{KITTI and Ford Campus:} Two training strategies are relevant for PoseNet + \textbf{F2M} but carry different meanings. 
The first is to use all available data for training, using the regularizing effect of the data to improve PoseNet prediction.
This is relevant when the LiDAR odometry is used offline, for example, as a mapping algorithm.
The second uses only data unobserved during training and evaluates the capacity of PoseNet to be used online as an initialization method. 
To evaluate both scenarios, we train PoseNet on two different splits of KITTI, sequences \texttt{00-21} for the mapping scenario and \texttt{11-21} for the online scenario. The results are presented in Table \ref{table_initialization}. 
They show that PoseNet improves \textbf{P-F2M} slightly, compared to the simpler \textbf{EI} and \textbf{CV} approaches, when running on data observed during training (see rows 1 and 3 on KITTI). 
However, this is not the case when PoseNet trained on unobserved data is considered, which leads to poorer performances.
We can make the connection with the previous Section : due to PoseNet's bad generalization properties, it is not yet well suited for inference, compared to simpler strategies, when either the motion or the environment differs from training. 
However, for offline mapping, using all available data for a specific configuration, allows small performance gains compared to simpler strategies. 

\begin{figure}
\figvspace
\centering
\includegraphics[width=0.80\linewidth]{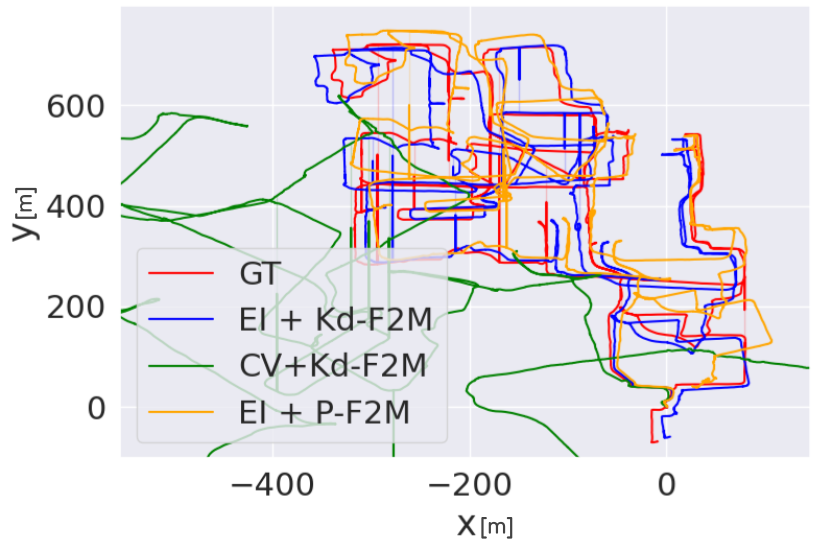}
\caption{Trajectories of sequence \texttt{01/08} of the NCLT dataset obtained by the proposed \textbf{Kd-F2M} for different initialization strategies. The figure shows that the constant velocity (\textbf{CV}) (green) model is less appropriate for this dataset than our 2D registration based estimator (\textbf{EI}). Ground Truth (GT) is in red.}
\label{michigan_initialization}
%\vspace{-5}
\end{figure}

\textbf{NCLT:} Classical LiDAR odometries in driving scenarios are very good precisely because the motion is easily predictable, leading to correct registration and in turn, high map quality. We now test our methods in the more challenging scenario of the NCLT dataset. The LiDAR mounted on the Segway is sparser (only 32 channels), and subject to abrupt rotations in yaw (around its axis) and pitch (the acceleration of the segway leads to an inclination of its support rod). We considered 10 sequences to train PoseNet in supervised and unsupervised settings (for more than 200000 frames in a repeating environment). The results on the sequences \texttt{01/08} (test) and \texttt{01/22} (train) for the proposed \textbf{F2M} with the different initialization strategies are presented in Table \ref{table_initialization_michigan} and Figure \ref{michigan_initialization}. 
First, they show poor performances for \textbf{CV} as an initialization method compared to \textbf{EI}. 
This illustrates well the importance of proper initialization for classical methods.
\textbf{EI} provides a good first estimate of the 2D motion (notably the rotation around the $z$ axis), which allows correct registration and thus, keeps the map clean. However the standard \textbf{CV} often fails to put the new scan in the convergence window of \textbf{F2M}, leading to maps of poor quality and thus, terrible pose registrations. 

Second, we can compare \textbf{P-F2M} and \textbf{Kd-F2M} in a more challenging registration setting. We see that our \textbf{Kd-F2M} behaves better than \textbf{P-F2M}. This is expected as the projective alignment leads to less precise neighbor associations. Furthermore, projective association is much more sensitive to abrupt rotations which can lead to large pixel distances between neighbors. However, the most important choice is the initialization strategy. Given an appropriate initialization strategy, even \textbf{P-F2M} provides correct results with reasonable drift as evidenced by Figure \ref{table_initialization_michigan}.

Finally, we see that PoseNet behaves poorly on the dataset in supervised and unsupervised settings, on the training and test sets. 
This demonstrates the limited representative power of our current PoseNet to more challenging motions. 
Despite using 10 times more data than for KITTI in a repeating environment, PoseNet is unable to learn a mapping from pairs of images to relative poses at a precision interesting enough to allow correct registrations. Due to the abrupt rotations, the motion of the sensor is located in a much greater volume of the parameters space than for driving motions. Thus, further research is required to allow neural network architectures to address these more challenging settings. \looseness=-1
%\vspace{-5}
% Test On MICHIGAN : potential of PoseNet over multiple trajectories

\begin{table}
\centering

\begin{tabular}{l|c|c} 
& Ford Campus & KITTI (\texttt{00-10}) \\
\toprule
\toprule
PoseNet \texttt{00-21}* + P-F2M & - & 0.66\\ \hline
PoseNet \texttt{11-21}* + P-F2M & - & 0.81\\ \hline
PoseNet Ford Campus* + P-F2M &  2.41 & - \\ \hline
CV + P-F2M & 2.04 & 0.69\\ \hline
EI + P-F2M & 2.11 & 0.69\\ 
\bottomrule
\end{tabular}
\caption{\tablefont Average Translation Error (ATE) on KITTI and Ford Campus for different initialization strategies, followed by \textbf{P-F2M} / \textbf{P-F2M}=Projective Frame-to-Model / CV="Constant Velocity", EI="Elevation Image"}
\label{table_initialization}
%\vspace{\dvs}
%\vspace{-10}
\end{table}

\begin{table}
\centering
%\vspace{-10}

\begin{tabular}{l|c|c } 
& \texttt{01/08} (test) & \texttt{01/22}* (train) \\
\toprule
\toprule
PoseNet (supervised) + Kd-F2M & 35.5 & 37.5 \\ \hline
PoseNet (unsupervised) + Kd-F2M & 48.7 & 46.3  \\ \hline
CV + Kd-F2M & 11.49 & 18.7 \\ \hline
EI + Kd-F2M & 1.84 & 3.73 \\ \hline 
EI + P-F2M & 4.32 & 5.21 \\ 
\bottomrule
\end{tabular}
\caption{\tablefont Average Translation Error on sequences \texttt{01/08} and \texttt{01/22} of the NCLT dataset, shows the importance of the selection of the initialization for LiDAR odometries}
\label{table_initialization_michigan}

\end{table}

%Compare PoseNet followed by Gauss Newton on with other initialization strategies, and mitigate the interest of using PoseNet as an initialization. 
%Initialization with training from one dataset to another, compared with GN + different motion models.
%Draw intresting conclusions.
 
\subsection{Weaknesses of Geometric LiDAR Odometries} \label{limits_geometric}

%MPrevious sections mostly present PoseNet weaknesses, and show that further research is required until it can compete with classical geometric methods.
%However PoseNet is mainly evaluated in situations where standard registration methods are already successful and very efficient. 
%Yet geometric LiDAR \textbf{F2M} odometries do fail, often, and without notice. 
%In this section, we search for such failure cases in our datasets, and show how PoseNet in an offline setting can increase the robustness of classical LiDAR odometries. 

Thus far, we have proven that registration-based \textbf{F2M} LiDAR odometries obtain superior results when a rich and dense map is constructed and when the new scan to be registered is moved into the small convergence domain of the mapping algorithm with good initialization.
The challenge is then the construction of a precise map and the prediction of the initial motion. 
This hints at the weaknesses of the classical methods: elements weakening the quality of the map (mobile objects, featureless regions, bad scan registration) as well as rapid and unpredictable motion (cf section \ref{analysis_initialization}). 

Another important weakness is the initial construction of the map. For the first two frames, \textbf{F2M} is equivalent to \textbf{F2F} without motion initialization. 
Bad initial map construction can be recovered for methods using a sliding local map (e.g. IMLS-SLAM \cite{imls_slam}), which can forget badly registered scans, but the trajectory and map constructed will only be correct locally, and not globally. 
It is much more problematic for methods relying on a global map (such as LOAM \cite{loam}).
We investigate this sensitivity in the driving datasets by counting the number of failures to align the first two initial frames.
Given a predicted pose $\mathbf{t}_{pred}=(\mathbf{t_x, t_y, t_z}),\mathbf{e}_{pred}=(\mathbf{e_x, e_y, e_z})$ and a ground truth pose $(\mathbf{t}_{gt}, \mathbf{e}_{gt})$ (using Euler representation for rotations), we categorize it as a registration failure if $||\mathbf{t}_{pred} - \mathbf{t}_{gt}||_2 > 1m$ or $|| \mathbf{e}_{pred}-\mathbf{e}_{gt}||_2>3^{\circ}$. 
The number of such failure cases for different \textbf{F2F} odometries over the Ford Campus and KITTI datasets is reported in Table \ref{table_failures}.
It confirms the known limitations of classical ICP-based \textbf{F2F} alignments, which have many such failures, especially when using projective neighbor association. 
Despite its simplicity, our \textbf{EI} initialization performs well on these outdoor datasets, with few failure cases. 
For this metric, PoseNet also performs really well both on training data (rows 1 and 3 of Table \ref{table_failures}), and on unobserved data (row 2).
Thus, despite its weak generalization properties (see section \ref{understanding}), PoseNet (when it can be trained) can provide more often a good enough initial estimate falling into the convergence domain of our \textbf{F2M}. 
Furthermore, the risk of absurd registrations is lower than for ICP-based \textbf{F2F}, as PoseNet essentially outputs poses observed during training, while ICP outside the convergence domain can potentially move in any direction of the parameter's space. 
In a sense, PoseNet encodes the motion model of the training dataset and can be used to increase the robustness of \textbf{F2M} LiDAR odometries, at the cost of offline training on similar data.

% GOAL OF THE SECTION : FIND VALID APPLICABLE USAGE OF POSENET / Expose the limitations of Classical methods 
%
% Violent unpredictable motions / Lack of geometric features (tunnel) / Mobile objects (which corrupt the map) /
% The main issue is that these problems are typically,  
% The wisdom to keep in mind : if PoseNet is able to learn, it must have a good enough supervision signal, 
% if there is a good enough supervision signal, then classical methods would probably work (however maybe requiring more engineering to suppress outliers and such) 
% 
% IF time : explore the failure cases on the dataset MICHIGAN (rapid rotation around axis, featureless environments, etc...)
% TODO : Explore elements of failure when considering on PandaSet and remove mobile objects etc...

\begin{table}
\centering
\figvspace
\begin{tabular}{l|c|c}
     & Ford Campus & KITTI (\texttt{00-10}) \\ \hline
\toprule
\toprule
  PoseNet (\texttt{00-21})*  & - & 0 \\ \hline
  PoseNet (\texttt{11-21})*  & - & 0 \\ \hline
  PoseNet (Ford Campus)*  & 261 & - \\ \hline
  NI + P-F2F  & 1494 &	3286 \\ \hline
  EI + P-F2F & 270 &  257 \\ \hline 
  NI + Kd-F2F & 1127 & 1845 \\ \hline
  EI + Kd-F2F & 261 & 0 \\
  \midrule 
  Number of Frames & 9996 & 23201 \\
  \bottomrule
\end{tabular}
\caption{\tablefont Number of failure over driving datasets}

\label{table_failures}
\end{table}

\section{Conclusion}

We analyzed, implemented, and compared classical and deep LiDAR odometry methods. 
We showed that the latter can overcome some of the former's weaknesses, but that further work is required before they can realistically be incorporated into practical odometry pipelines. 
Our analysis allowed us to build a modular architecture, which we make available in {\bf \texttt{pyLIDAR-SLAM}} and intend to extend and improve in future works, aiming to integrate more closely deep and classical LiDAR odometries.

%\small
\bibliographystyle{IEEEtran}
\bibliography{IEEEabrv,biblio}

\end{document}